\documentclass[10pt,twocolumn,letterpaper]{article}

\usepackage{cvpr}
\usepackage{times}
\usepackage{epsfig}
\usepackage{graphicx}
\usepackage{amsmath}
\usepackage{amssymb}

\usepackage{algorithm}
\usepackage{algorithmicx}
\usepackage{algpseudocode}
\usepackage{color}
\usepackage{enumitem}
\usepackage{arydshln}
\usepackage{multirow}
\usepackage{bm}

\usepackage[breaklinks=true,bookmarks=false]{hyperref}

\cvprfinalcopy 

\def\cvprPaperID{****} 

\begin{document}

\title{Shape Robust Text Detection with Progressive Scale Expansion Network}

\author{Wenhai Wang$^{1,4*}$, Enze Xie$^{2,5*}$, Xiang Li$^{3,4*}$\thanks{Authors contributed equally.}\thanks{Xiang Li is with PCA Lab, Key Lab of Intelligent Perception and Systems for High-Dimensional Information of Ministry of Education, and Jiangsu Key Lab of Image and Video Understanding for Social Security, School of Computer Science and Engineering, Nanjing University of Science and Technology, Nanjing, 210094, China. Xiang Li is also a visiting scholar in Momenta.},  Wenbo Hou$^{1}$, Tong Lu$^{1}$\thanks{Corresponding author.}, Gang Yu$^{5}$, Shuai Shao$^{5}$\\
${^1}$National Key Lab for Novel Software Technology, Nanjing University\\
${^2}$Department of Comuter Science and Technology, Tongji University\\
${^3}$School of Computer and Engineering, Nanjing University of Science and Technology\\
${^4}$Momenta \\
${^5}$Megvii (Face++) Technology Inc.\\
\tt\small\{wangwenhai362, Johnny\_ez, lysucuo\}@163.com, xiang.li.implus@qq.com\\
\tt\small lutong@nju.edu.cn, \{yugang, shaoshuai\}@megvii.com
}

\maketitle

\begin{abstract}
   Scene text detection has witnessed rapid progress especially with the recent development of convolutional neural networks. However, there still exists two challenges which prevent the algorithm into industry applications. On the one hand, most of the state-of-art algorithms require quadrangle bounding box which is in-accurate to locate the texts with arbitrary shape. On the other hand, two text instances which are close to each other may lead to a false detection which covers both instances. Traditionally, the segmentation-based approach can relieve the first problem but usually fail to solve the second challenge. To address these two challenges, in this paper, we propose a novel Progressive Scale Expansion Network (PSENet), which can precisely detect text instances with arbitrary shapes. More specifically, PSENet generates the different scale of kernels for each text instance, and gradually expands the minimal scale kernel to the text instance with the complete shape. Due to the fact that there are large geometrical margins among the minimal scale kernels, our method is effective to split the close text instances, making it easier to use segmentation-based methods to detect arbitrary-shaped text instances. Extensive experiments on CTW1500, Total-Text, ICDAR 2015 and ICDAR 2017 MLT validate the effectiveness of PSENet. Notably, on CTW1500, a dataset full of long curve texts, PSENet achieves a F-measure of 74.3\% at 27 FPS, and our best F-measure (82.2\%) outperforms state-of-art algorithms by 6.6\%. 
	The code will be released in the future.
\end{abstract}

\begin{figure}
	\centering
	\setlength{\fboxrule}{0pt}
	\fbox{\includegraphics[width=0.4\textwidth]{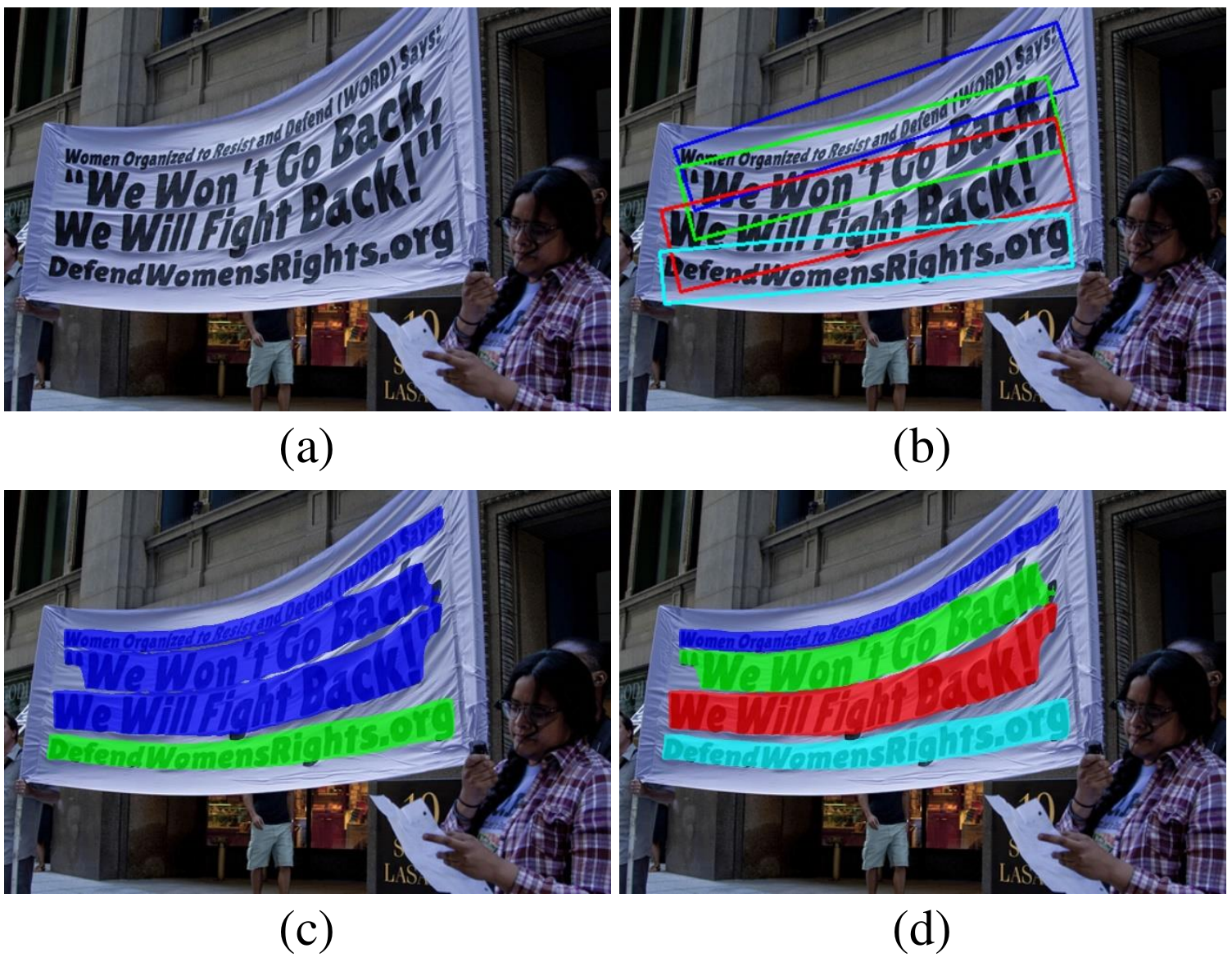}}
	\caption{The results of different methods, best viewed in color. (a) is the original image. (b) refers to the result of regression-based method, which displays disappointing detections as the red box covers nearly more than half of the context in the green box. (c) is the result of naive semantic segmentation, which mistakes 3 text instances as 1 instance since their boundary pixels are partially connected. (d) is the result of our proposed PSENet, which successfully distinguishs and detects the 4 unique text instances.}
	\label{fig:diff-meth-res}
\end{figure}
\section{Introduction}
	Scene text detection in the wild is a fundamental problem with numerous applications such as scene understanding, product identification, and autonomous driving. Many progress has been made in recent years with the rapid development of Convolutional Neural Networks (CNNs)~\cite{he2016deep,huang2017densely,ren2015faster}. We can roughly divide the existing CNN based algorithm into two categories: regression-based approaches and segmentation-based approaches.
	
	For the regression-based approaches~\cite{tian2016detecting, zhou2017east, shi2017detecting, jiang2017r2cnn, zhong2016deeptext, liu2018fots, he2017single, hu2017wordsup, lyu2018multi}, the text targets are usually represented in the forms of rectangles or quadrangles with certain orientations. However, the regression-based approaches fail to deal with the text instance with arbitrary shapes, e.g., the curve texts as shown in Fig.~\ref{fig:diff-meth-res}~(b). Segmentation-based approaches, on the other hand, locate the text instance based on pixel-level classification. However, it is difficult to separate the text instances which are close with each other. Usually, a false detection which covers all the text instances close to each other may be predicted based on the segmentation-based approach. One example is shown in Fig.~\ref{fig:diff-meth-res}~(c).
	
	To address these problems, in this paper, we propose a novel kernel-based framework, namely, Progressive Scale Expansion Network (PSENet). Our PSENet has the following two benefits. First, as a segmentation-based method, PSENet performs pixel-level segmentation, which is able to precisely locate the text instance with arbitrary shape. Second, we propose a progressive scale expansion algorithm, with which the adjacent text instances can be successfully identified as shown in Fig.~\ref{fig:diff-meth-res}~(d). More specifically, we assign each text instance with multiple predicted segmentation areas, which are denoted as ``kernels'' for simplicity.  Each kernel has the similar shape to the original text instance but different scales. To obtain the final detections, we adopt a progressive scale expansion algorithm based on Breadth-First-Search (BFS). Generally, there are 3 steps: 1) starting from the kernels with minimal scales (instances can be distinguished in this step); 2) expanding their areas by involving more pixels in larger kernels gradually; 3) finishing until the complete text instances (the largest kernels) are explored.
	
\begin{figure}
		\centering
		\setlength{\fboxrule}{0pt}
		\fbox{\includegraphics[width=0.4\textwidth]{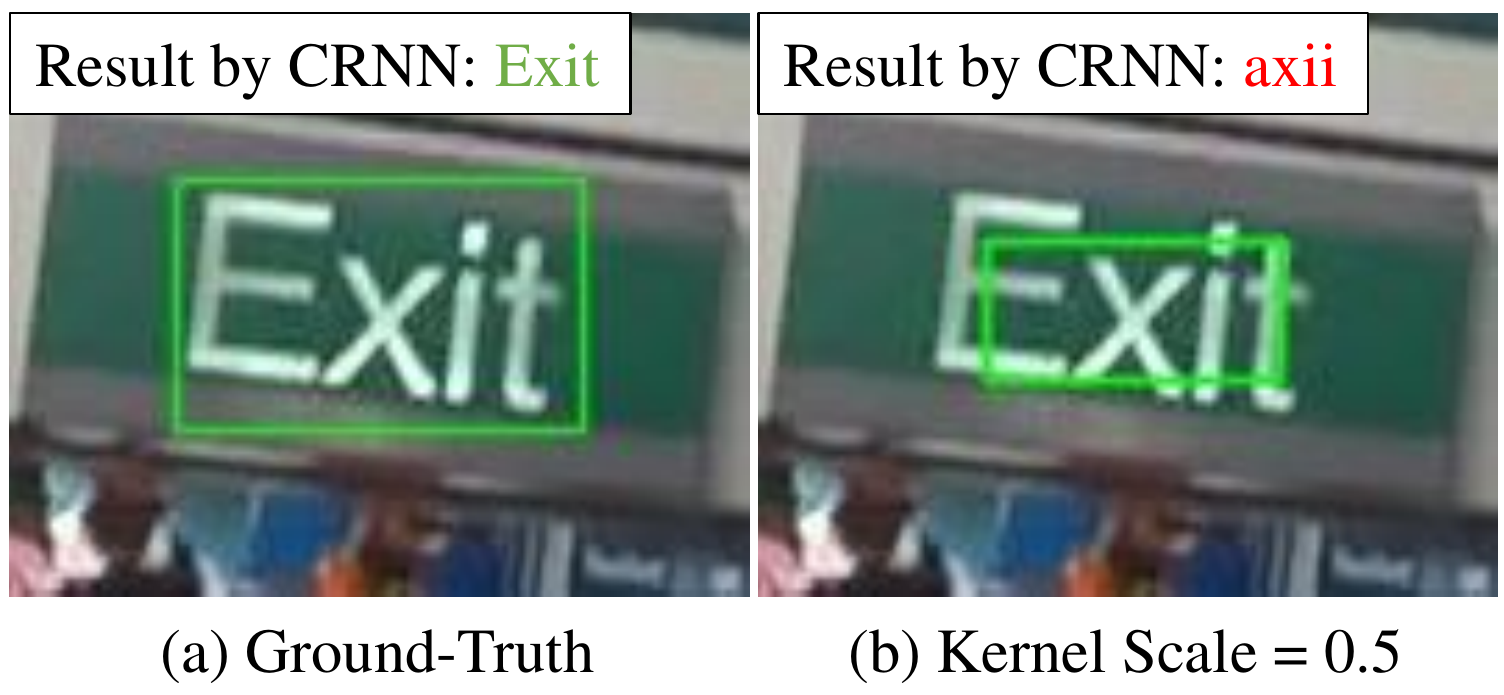}}
		\caption{Visualization of complete text instance and kernel of text instance. It can be seen that CRNN~\cite{crnn} recognizes complete text instance correctly but fail to recognize the kernel, because the kernel can not cover the complete areas of text instances.}
		\label{fig:kernel_scale}
	\end{figure}

	There are three potential reasons for the design of the progressive scale expansion algorithm. First, the kernels with minimal scales are quite easy to be separated as their boundaries are far away from each other. Second, the minimal scale kernels can not cover the complete areas of text instances (see Fig.~\ref{fig:kernel_scale}~(b)). Therefore, it is necessary to recover the complete text instances from the minimal scale kernels. Third, the progressive scale expansion algorithm is a simple and efficient method to expand the small kernels to complete text instances, which ensures the accurate locations of text instances.
	
	To show the effectiveness of our proposed PSENet, we conduct extensive experiments on four competitive benchmark datasets including ICDAR 2015~\cite{karatzas2015icdar}, ICDAR 2017 MLT~\cite{icdar2017mlt} ,CTW1500~\cite{Liu2017Detecting} and Total-Text~\cite{totaltext}. Among these datasets, CTW1500 and Total-Text are explicitly designed for curve text detection. 
	Specifically, on CTW1500, a dataset with long curve texts, we outperform state-of-the-art results by absolute 6.6\%, and our real-time model achieves a comparable performance (74.3\%) at 27 FPS. Furthermore, the proposed PSENet also achieves promising performance on multi-oriented and multi-lingual text datasets: ICDAR 2015 and ICDAR 2017 MLT.

	\begin{figure*}[t]
		\centering
		\setlength{\fboxrule}{0pt}
		\fbox{\includegraphics[width=0.85\textwidth]{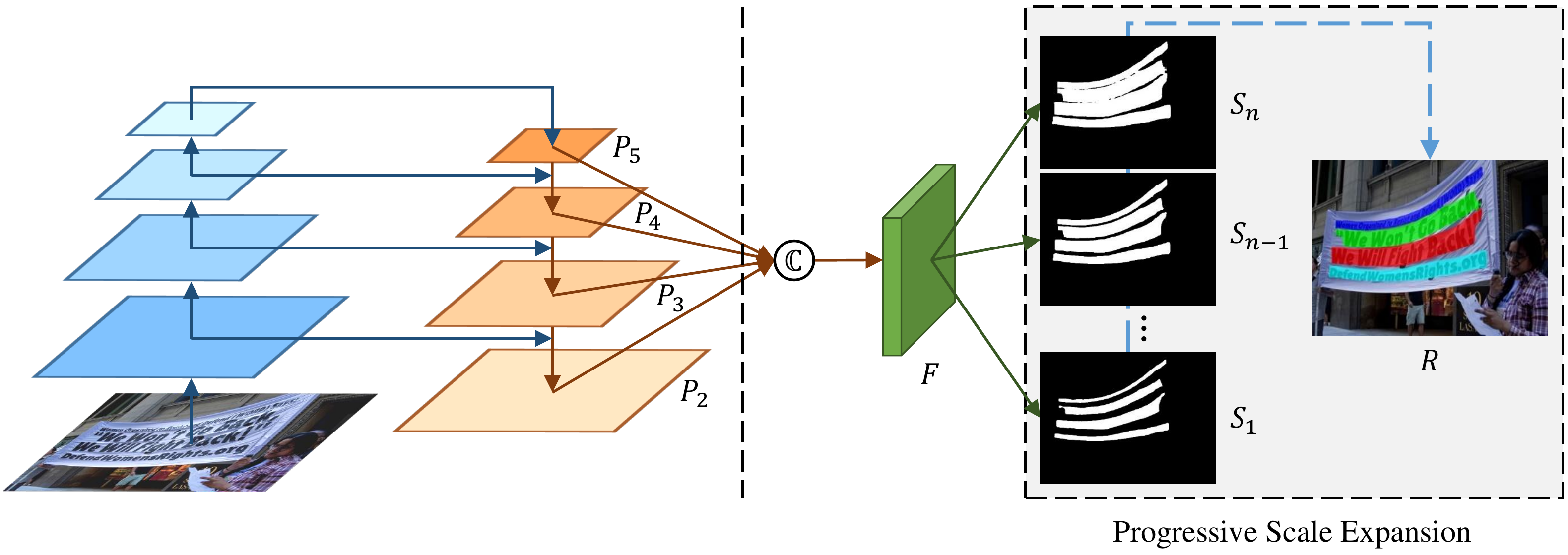}}
		\caption{Illustration of our overall pipeline. The left part of pipeline is implemented from FPN~\cite{lin2017feature}. The right part denotes the feature fusion and the progressive scale expansion algorithm.}
		\label{fig:pipeline}
	\end{figure*}
	
	\section{Related Work}
	Scene text detection based on deep learning methods have achieved remarkable results over the past few years. A major of modern text detectors are based on CNN framework, in which scene text detection is roughly formulated as two categories: regression-based methods and segmentation-based methods.
	
	\textbf{Regression-based methods} often based on general object detection frameworks, such Faster R-CNN~\cite{ren2015faster} and SSD~\cite{liu2016ssd}.
	TextBoxes~\cite{liao2017textboxes} modified the anchor scales and shape of convolution kernels to adjust to the various aspect ratios of the text.
	EAST~\cite{zhou2017east} use FCN~\cite{FCN} to directly predict score map, rotation angle and text boxes for each pixel.
	RRPN~\cite{rrpn} adopted Faster R-CNN and developed rotation proposals of RPN part to detect arbitrary oriented text. 
	RRD~\cite{rrd} extracted feature maps for text classification and regression from two separately branches to better long text detection.
	
	However, most of the regression-based methods often require complex anchor design and cumbersome multiple stages, which might require exhaustive tuning and lead to sub-optimal performance. Moreover, the above works were specially designed for multiple oriented text detection and may fall short when handling curve texts, which are actually widely distributed in
	real-world scenarios.
	
	\textbf{Segmentation-based methods} are mainly inspired by fully convolutional networks(FCN)~\cite{FCN}. 
	Zhang $\it{et~al.}$~\cite{zhang2016multi} first adopted FCN to extract text blocks and detect character candidates from those text blocks via MSER.
	Yao $\it{et~al.}$~\cite{yao2016scene} formulated one text region as various
	properties, such as text region and orientation, then utilized FCN to predict the corresponding heatmaps.
	Lyu $\it{et~al.}$\cite{lyu2018multi} utilized corner localization to find suitable irregular quadrangles for text instances. 
	PixelLink~\cite{PixelLink} separated texts lying close to each other by predicting pixel connections between different text instances.
	Recently, TextSnake~\cite{textsnake} used ordered disks to represent curve text for curve text detection.
	SPCNet~\cite{xie2018scene} used instance segmentation framework and utilize context information to detect text of arbitrary shape while suppressing false positives.
	
	The above works have achieved excellent performances over several horizontal and multi-oriented text benchmarks.
	Similarly, most of the above approaches have not paid special attention to curve text, except for TextSnake~\cite{textsnake}. 
	However, TextSnake still needs time-consuming and complicated
	post-processing steps~(Centralizing, Striding and Sliding) during inference, while our proposed Progressive Scale Expansion needs only one clean and efficient step.
	
	
	\section{Proposed Method}
	In this section, we first introduce the overall pipeline of the proposed Progressive Scale Expansion Network (PSENet). Next, we present the details of progressive scale expansion algorithm, and show how it can effectively distinguish the text instances lying closely. At last, we introduce the way of generating label and the design of loss function.
	
	\subsection{Overall Pipeline}
	A high-level overview of our proposed PSENet is illustrated in Fig.~\ref{fig:pipeline}. 
	We use ResNet~\cite{he2016identity} as the backbone of PSENet. 
	We concatenate low-level texture feature with high-level semantic feature. These maps are further fused in $F$ to encode information with various receptive views. Intuitively, such fusion is very likely to facilitate the generations of the kernels with various scales. Then the feature map $F$ is projected into $n$ branches to produce multiple segmentation results $S_1, S_2, ..., S_n$. Each $S_i$ would be one segmentation mask for all the text instances at a certain scale. The scales of different segmentation mask are decided by the hyper-parameters which will be discussed in Sec.~\ref{sec:label-gen}.
	Among these masks, $S_1$ gives the segmentation result for the text instances with smallest scales (i.e., the minimal kernels) and $S_n$ denotes for the original segmentation mask (i.e., the maximal kernels). After obtaining these segmentation masks, we use progressive scale expansion algorithm to gradually expand all the instances' kernels in $S_1$, to their complete shapes in $S_n$, and obtain the final detection results as $R$.
	
	\subsection{Network Design}
	The basic framework of PSENet is implemented from FPN~\cite{lin2017feature}. We firstly get four $256$ channels feature maps (i.e. $P_2, P_3, P_4, P_5$) from the backbone. To further combine the semantic features from low to high levels, we fuse the four feature maps to get feature map $F$ with $1024$ channels via the function $\mathbb{C}(\cdot)$ as:
	\begin{equation}
	\begin{split}
	F &= \mathbb{C}(P_2, P_3, P_4, P_5) \\
	&= P_2 \parallel \mbox{Up}_{\times 2}(P_3) \parallel \mbox{Up}_{\times 4}(P_4) \parallel \mbox{Up}_{\times 8}(P_5),
	\label{eqn:c_fun}
	\end{split}
	\end{equation}
	where ``$\parallel$'' refers to the concatenation and $\mbox{Up}_{\times 2}(\cdot)$, $\mbox{Up}_{\times 4}(\cdot)$, $\mbox{Up}_{\times 8}(\cdot)$ refer to 2, 4, 8 times upsampling, respectively. Subsequently, $F$ is fed into Conv($3, 3$)-BN-ReLU layers and is reduced to $256$ channels. Next, it passes through $n$ 
	Conv($1, 1$)-$\mbox{Up}$-Sigmoid layers and produces $n$ segmentation results $S_1, S_2, ..., S_n$. Here, Conv, BN, ReLU and $\mbox{Up}$ refer to convolution~\cite{lecun1998gradient}, batch normalization~\cite{ioffe2015batch}, rectified linear units~\cite{glorot2011deep} and upsampling.
	
	\begin{figure*}
		\centering
		\setlength{\fboxrule}{0pt}
		\fbox{\includegraphics[width=0.9\textwidth]{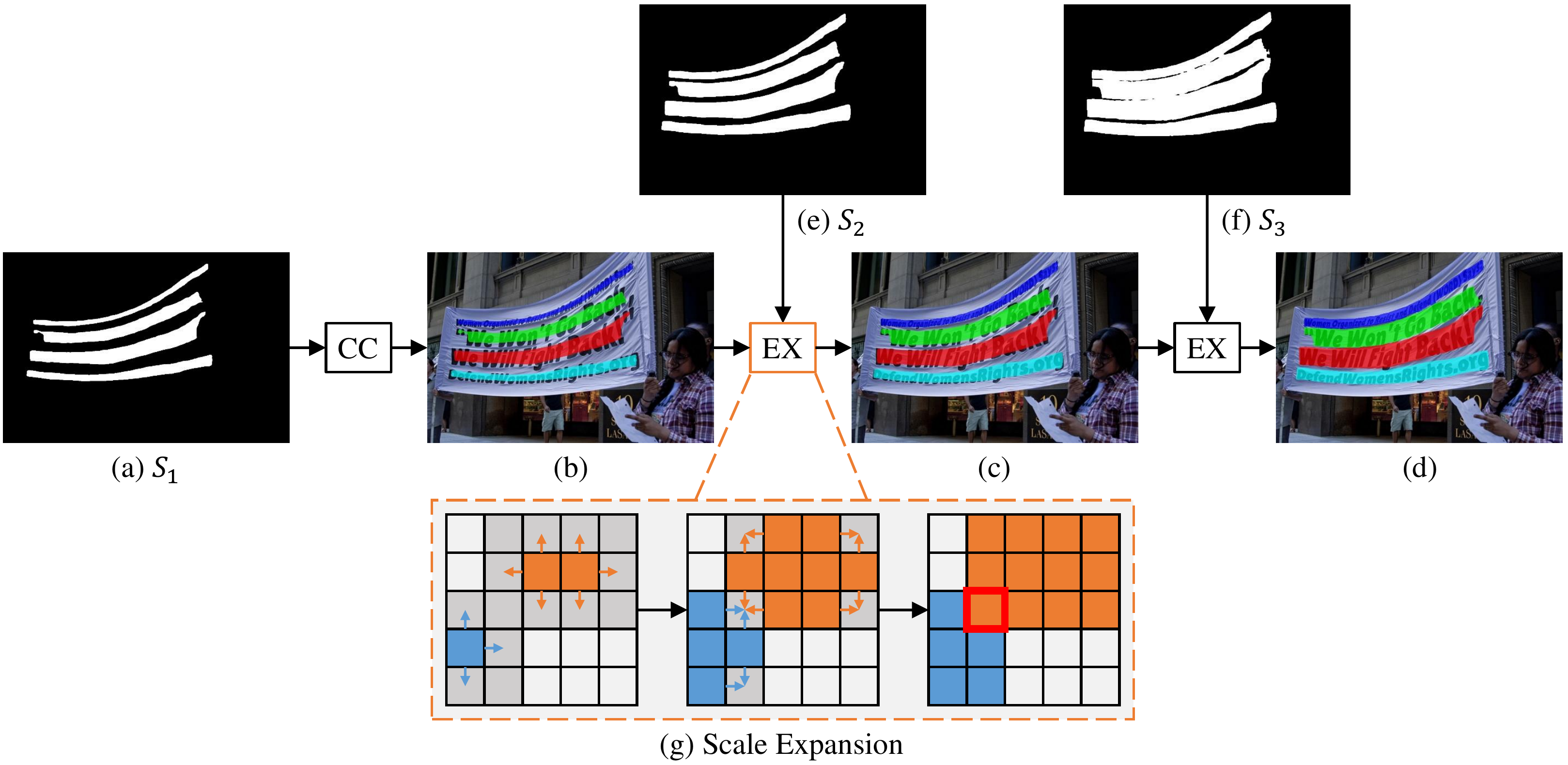}}
		\caption{The procedure of progressive scale expansion algorithm. CC refers to the function of finding connected components. EX represents the scale expansion algorithm. (a), (e) and (f) refer to $S_1$, $S_2$ and $S_3$, respectively. (b) is the initial connected components. (c) and (d) is the results of expansion. (g) is the illustration of expansion. The blue and orange areas represent the kernels of different text instances. The gray girds represent the pixels need to be involved. The red box in (g) refers to the conflicted pixel.}
		\label{fig:pse}
	\end{figure*}
	
	\subsection{Progressive Scale Expansion Algorithm}
	As shown in Fig.~\ref{fig:diff-meth-res}~(c), it is hard for the segmentation-based method to separate the text instances that are close to each other. To solve this problem, we propose a progressive scale expansion algorithm. 
	
	
	Here is a vivid example (see Fig.~\ref{fig:pse}) to explain the procedure of progressive scale expansion algorithm, whose central idea is brought from the Breadth-First-Search (BFS) algorithm. In the example, we have $3$ segmentation results $S = \{S_1, S_2, S_3\}$ (see Fig.~\ref{fig:pse}~(a),~(e),~(f)). At first, based on the minimal kernels' map $S_1$ (see Fig.~\ref{fig:pse}~(a)), 4 distinct connected components $C = \{c_1, c_2, c_3, c_4\}$ can be found as initializations. The regions with different colors in Fig.~\ref{fig:pse} (b) represent these different connected components, respectively. By now we have all the text instances' central parts (i.e., the minimal kernels) detected. 
	Then, we progressively expand the detected kernels by merging the pixels in $S_2$, and then in $S_3$. The results of the two scale expansions are shown in Fig.~\ref{fig:pse}~(c) and Fig.~\ref{fig:pse}~(d), respectively.
	Finally, we extract the connected components which are marked with different colors in Fig.~\ref{fig:pse}~(d) as the final predictions for text instances.
	
	
	The procedure of scale expansion is illustrated in Fig.~\ref{fig:pse}~(g). The expansion is based on Breadth-First-Search algorithm which starts from the pixels of multiple kernels and iteratively merges the adjacent text pixels. Note that there may be conflicted pixels during expansion, as shown in the red box in Fig.~\ref{fig:pse}~(g). The principle to deal with the conflict in our practice is that the confusing pixel can only be merged by one single kernel on a first-come-first-served basis. Thanks to the ``progressive'' expansion procedure, these boundary conflicts will not affect the final detections and the performances.
	The detail of scale expansion algorithm is summarized in Algorithm~\ref{alg:expansion}. In the pseudocode, $T, P$ are the intermediate results. $Q$ is a queue. $\mbox{Neighbor}(\cdot)$ represents the neighbor pixels (4-ways) of $p$. $\mbox{GroupByLabel}(\cdot)$ is the function of grouping the intermediate result by label. ``$S_i[q] = \mbox{True}$'' means that the predicted value of pixel $q$ in $S_i$ belongs to the text part. $C$ and $E$ are used to keep the kernels before and after expansion
	respectively;
	
	\begin{algorithm}[t]
		\scriptsize
		\caption{Scale Expansion Algorithm}
		\begin{algorithmic}[1]
			\Require Kernels: $C$, Segmentation Result: $S_i$
			\Ensure Scale Expanded Kernels: $E$
			\Function {Expansion}{$C$, $S_i$}
			\State $T \gets \emptyset; P \gets \emptyset; Q \gets \emptyset$
			\For{each $c_i \in C$}
			\State $T \gets T \cup \{(p, label) \mid (p, label) \in c_i\}$
			\State $P \gets P \cup \{p \mid (p, label) \in c_i\}$
			\State $\mbox{\bf{Enqueue}}(Q, c_i)$ \ \ \ \ \ \ \ \ \ \ \ \ \ \ \ \ \ \ \ \ \ \ \ \ \ \ \ \ // push all the elements in $c_i$ into $Q$
			\EndFor
			\While{$Q \neq \emptyset$}
			\State $(p, label) \gets \mbox{\bf{Dequeue}}(Q)$   \ \ \ \ \ \ \ \ \ \ // pop the first element of $Q$
			\If{$\exists q \in \mbox{\bf{Neighbor}}(p)$ and $q \notin P$ and $S_i[q] = \mbox{True}$}
			\State $T \gets T \cup \{(q, label)\}; P \gets P \cup \{q\}$
			\State $\mbox{\bf{Enqueue}}(Q, (q, label))$ \ \ \ \ \ \ \ // push the element $(q, label)$ into $Q$
			\EndIf
			\EndWhile
			\State $E = \mbox{\bf{GroupByLabel}}(T)$
			\State \Return{$E$}
			\EndFunction
		\end{algorithmic}
		\label{alg:expansion}
	\end{algorithm}
	
	\subsection{Label Generation}
	\label{sec:label-gen}
\begin{figure}[t]
		\centering
		\setlength{\fboxrule}{0pt}
		\fbox{\includegraphics[width=0.45\textwidth]{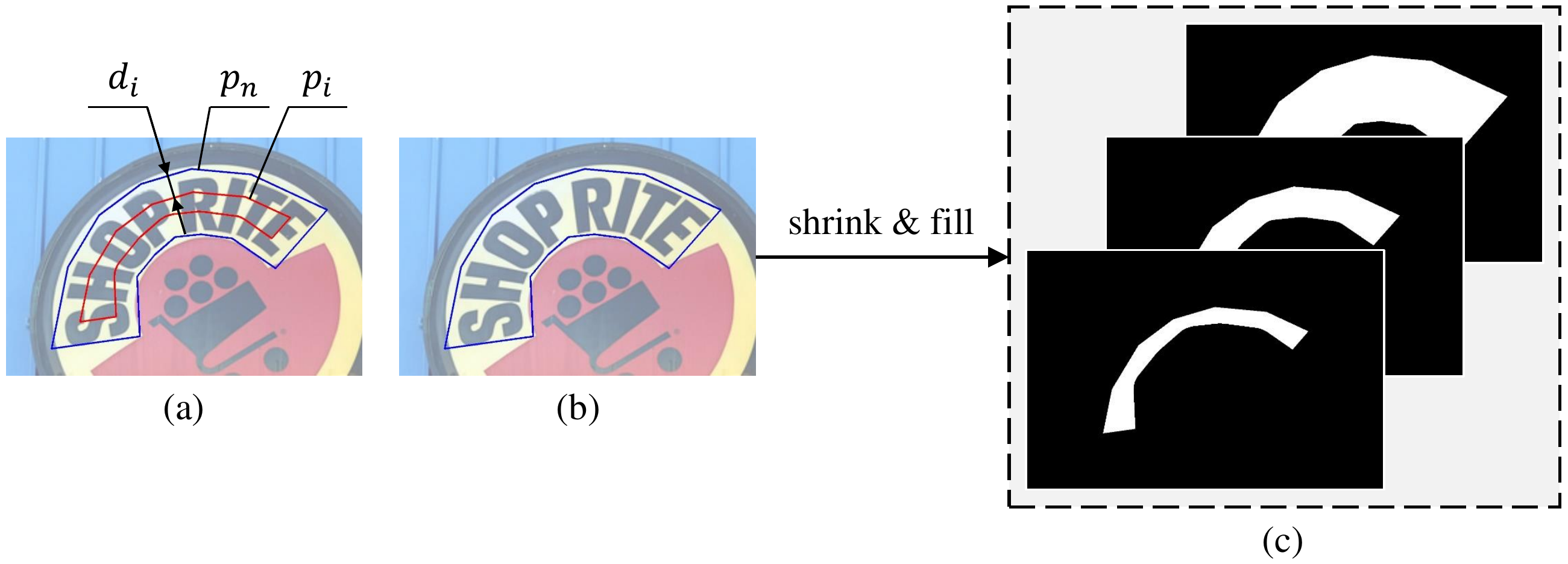}}
		\caption{The illustration of label generation. (a) contains the annotations for $d$, $p_i$ and $p_n$. (b) shows the original text instances. (c) shows the segmentation masks with different kernel scales.}
		\label{fig:label_gen}
		
	\end{figure}

	As illustrated in Fig.~\ref{fig:pipeline}, PSENet produces segmentation results (e.g. $S_1, S_2, ..., S_n$) with different kernel scales. Therefore, it requires the corresponding ground truths with different kernel scales during training. In our practice, these ground truth labels can be conducted simply and effectively by shrinking the original text instance.
	The polygon with blue border in Fig.~\ref{fig:label_gen}~(b) denotes the original text instance 
	and it corresponds to the largest segmentation label mask (see the rightmost map in Fig.~\ref{fig:label_gen}~(c)). To obtain the shrunk masks sequentially in Fig.~\ref{fig:label_gen}~(c), we utilize the Vatti clipping algorithm~\cite{vatti1992generic} to shrink the original polygon $p_n$ by $d_i$ pixels and get shrunk polygon $p_i$ (see Fig.~\ref{fig:label_gen}~(a)). Subsequently, each shrunk polygon $p_i$ is transferred into a 0/1 binary mask for segmentation label ground truth. We denote these ground truth maps as $G_1, G_2, ..., G_n$ respectively. Mathematically, if we consider the scale ratio as $r_i$, the margin $d_i$ between $p_n$ and $p_i$ can be calculated as:
	\begin{equation}
	d_i = \frac{\mbox{Area}(p_n) \times (1 - r_i^2)}{\mbox{Perimeter}(p_n)},
	\label{eqn:d_i}
	\end{equation}
	where $\mbox{Area}(\cdot)$ is the function of computing the polygon area, $\mbox{Perimeter}(\cdot)$ is the function of computing the polygon perimeter. Further, we define the scale ratio $r_i$ for ground truth map $G_i$ as:
	\begin{equation}
	r_i = 1 - \frac{(1 - m) \times (n - i)}{n - 1},
	\label{eqn:r_i}
	\end{equation}
	where $m$ is the minimal scale ratio, which is a value in $(0, 1]$. Based on the definition in Eqn.~\eqref{eqn:r_i}, the values of scale ratios (i.e., $r_1, r_2, ..., r_n$) are decided by two hyper-parameters $n$ and $m$, and they increase linearly from $m$ to $1$.

	\subsection{Loss Function}
	For learning PSENet, the loss function can be formulated as:
	\begin{equation}
	L =  \lambda L_c + (1 - \lambda) L_s,
	\label{eqn:loss-tot}
	\end{equation}
	where $L_c$ and $L_s$ represent the losses for the complete text instances and the shrunk ones respectively, and $\lambda$ balances the importance between $L_c$ and $L_s$. 
	
	It is common that the text instances usually occupy only an extremely small region in natural images, which makes the predictions of network bias to the non-text region, when binary cross entropy~\cite{de2005tutorial} is used. Inspired by~\cite{milletari2016v}, we adopt dice coefficient in our experiment. The dice coefficient $D(S_i, G_i)$ is formulated as in Eqn.~\eqref{eqn:dice-coef}:
	\begin{equation}
	D(S_i, G_i) = \frac{2 \sum\nolimits_{x,y} (S_{i, x, y} \times G_{i, x, y})}{\sum\nolimits_{x, y} S_{i, x, y}^2 + \sum\nolimits_{x, y} G_{i, x, y}^2},
	\label{eqn:dice-coef}
	\end{equation}
	where $S_{i, x, y}$ and $G_{i, x, y}$ refer to the value of pixel $(x, y)$ in segmentation result $S_i$ and ground truth $G_i$, respectively. 
	
	Furthermore, there are many patterns similar to text strokes, such as fences, lattices, etc. Therefore, we adopt Online Hard Example Mining (OHEM)~\cite{shrivastava2016training} to $L_c$ during training to better distinguish these patterns.
	
	$L_c$ focuses on segmenting the text and non-text region. Let us consider the training mask given by OHEM as $M$, and thus $L_c$ can be formulated as Eqn.~\eqref{eqn:loss-complete}.
	\begin{equation}
	L_c = 1 - D(S_n \cdot M, G_n \cdot M),
	\label{eqn:loss-complete}
	\end{equation}
	$L_s$ is the loss for shrunk text instances. Since they are encircled by the original areas of the complete text instances, we ignore the pixels of the non-text region in the segmentation result $S_n$ to avoid a certain redundancy. Therefore, $L_s$ can be formulated as follows:
	\begin{equation}
	\begin{split}
	L_s = 1 - \frac{\sum\nolimits_{i = 1}^{n - 1} D(S_i \cdot W, G_i \cdot W)}{n - 1},\\
	W_{x, y} = 
	\left\{
	\begin{array}{ll}
	1, & if \ S_{n, x, y} \ge 0.5; \\  
	0, & otherwise.   
	\end{array}
	\right.  
	\label{eqn:loss-shrink}
	\end{split}
	\end{equation}
	
	Here, $W$ is a mask which ignores the pixels of the non-text region in $S_n$, and $S_{n, x, y}$ refers to the value of pixel $(x, y)$ in $S_n$.
	
	\section{Experiment}
	In this section, we first briefly introduce datasets and present the details of implementation.
	Then, we conduct ablation studies for PSENet. At last, we evaluate the proposed PSENet on four recent challenging public benchmarks: CTW1500, Total-Text, ICDAR 2015 and ICDAR 2017 MLT, and compare PSENet with state-of-the-art methods.
	
	\subsection{Datasets}
	
	\textbf{CTW1500}~\cite{Liu2017Detecting} is a challenging dataset for long curve text detection, which is constructed by Yuliang et al.~\cite{Liu2017Detecting}. It consists of 1000 training images and 500 testing images. Different from traditional text datasets (e.g. ICDAR 2015, ICDAR 2017 MLT), the text instances in CTW1500 are labelled by a polygon with 14 points which can describe the shape of an arbitrarily curve text.
	
	\textbf{Total-Text}~\cite{totaltext} is a newly-released dataset for curve text detection. Horizontal, multi-Oriented and curve text instances are contained in Total-Text. The benchmark consists of 1255 training images and 300 testing images.
	
	\textbf{ICDAR 2015} (IC15)~\cite{karatzas2015icdar} is a commonly used dataset for text detection. It contains a total of 1500 pictures, 1000 of which are used for training and the remaining are for testing. The text regions are annotated by 4 vertices of the quadrangle.
	
	\textbf{ICDAR 2017 MLT} (IC17-MLT)~\cite{icdar2017mlt} is a large scale multi-lingual text dataset, which includes 7200 training images, 1800 validation images and 9000 testing images. The dataset is composed of complete scene images which come from 9 languages. 
	
	\subsection{Implementation Details}
	We use the ResNet~\cite{he2016identity} pre-trained on ImageNet~\cite{deng2009imagenet} as our backbone. 
	All the networks are optimized by using stochastic gradient descent (SGD).
	We use 7200 IC17-MLT training images and 1800 IC17-MLT validation images to train the model and report the result on IC17-MLT. Note that no extra data, e.g. SynthText~\cite{synthtext}, is adopted to train IC17-MLT.
	We train PSENet on IC17-MLT with batch size 16 on 4 GPUs for 180K iterations.
	The initial learning rate is set to $1\times10^{-3}$, and is divided by 10 at 60K and 120K iterations.
	 
	Two training strategies are adopted in the rest of all datasets:(1) Training from scratch. (2) Fine-tuning on IC17-MLT model. 
	When training from scratch, we train PSENet with batch size 16 on 4 GPUs for 36K iterations, and the initial learning rate is set to $1\times10^{-3}$ and is divided by 10 at 12K and 24K iterations.
	When fine-tuning on IC17-MLT model, the number of iterations is 24K, and the initial learning rate is $1\times10^{-4}$ which is divided by 10 at 12K iterations.
	
	We use a weight decay of $5 \times 10^{-4}$ and a Nesterov momentum~\cite{sutskever2013importance} of 0.99 without dampening. We adopt the weight initialization introduced by~\cite{he2015delving}.
	
	During training, we ignore the blurred text regions labeled as “DO NOT CARE” in all datasets. The $\lambda$ of loss balance is set to $0.7$. The negative-positive ratio of OHEM is set to 3. The data augmentation for training data is listed as follows: 
	1) the images are rescaled with ratio $\{0.5, 1.0, 2.0, 3.0\}$ randomly; 
	2) the images are horizontally flipped and rotated in the range $[-10^\circ, 10^\circ]$ randomly; 
	3) $640 \times 640$ random samples are cropped from the transformed images.
	For quadrangular text, we calculate the minimal area rectangle to extract the bounding boxes. For curve text dataset, the output of PSE is applied to produce the final result. 
	

	\subsection{Ablation Study}
	
	\begin{figure}[b]
		\centering
		\setlength{\fboxrule}{0pt}
		\fbox{\includegraphics[width=0.45\textwidth]{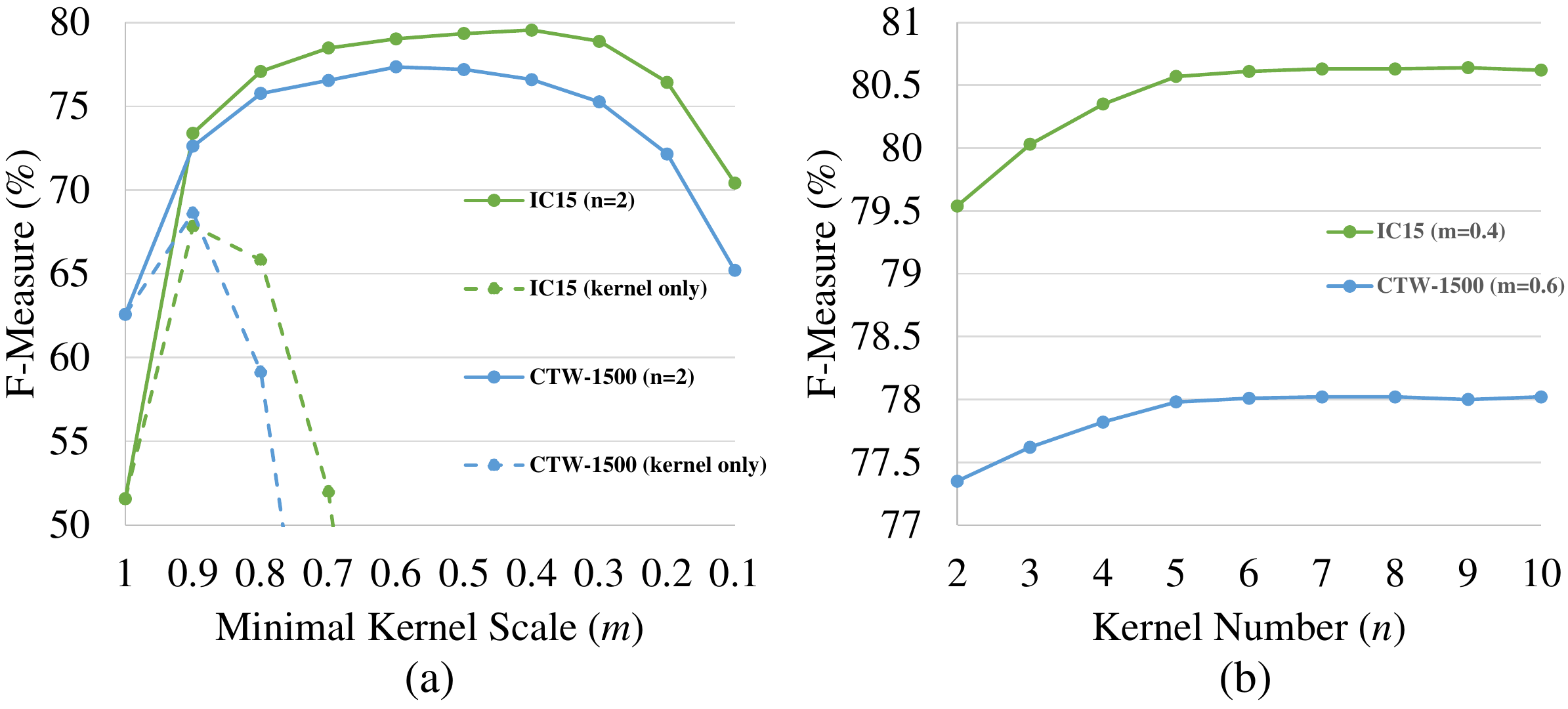}}
		\caption{Ablation study on minimal kernel scale ($m$) and kernel number ($n$)  (Eqn.~\eqref{eqn:r_i}). There results are based on PSENet-1s (Resnet 50) trained from scratch. ``1s'' means the shape of output map is $1/1$ of the input image. }
		\label{fig:abs}
	\end{figure}
	
	\textbf{Can kernels be used as the final result?} The aim of kernels is to roughly locate the text instance and separate the text instances standing closely to each other. However, the minimal scale kernels can not cover the complete areas of text instances, which does harm to the text detection and recognition. In Fig.~\ref{fig:abs}~(a), the F-measures of the detector used minimal scale kernel only (the dash curves) is terrible on ICDAR 2015 and CTW1500 datasets. In addition, we use a modern text recognizer CRNN~\cite{crnn} to recognize the text in complete text instance and kernel, and find that CRNN failed to recognize the text in the kernel (see Fig.~\ref{fig:kernel_scale}). Thus, the kernel can not be used as the final detection result.
	
	\textbf{Influence of the minimal kernel scale.} We study the effect of the minimal scale $m$ by setting the number of kernels $n$ to $2$ and let the minimal scale $m$ vary from $1$ to $0.1$. The models are evaluated on ICDAR 2015 and CTW1500 two datasets. We can find from Fig.~\ref{fig:abs}~(a) that the F-measures on the test sets drops when $m$ is too large or too small. 
	Note that when setting kernel scale 1, we only use text segmentation map as the final result and without progressive scale expansion algorithm. Obviously, without PSE the baseline's performance is unsatisfactory because the network fails to separate the text lying closely to each other.
	When $m$ is too large, it is hard for PSENet to separate the text instances lying closely to each other. When $m$ is too small, PSENet often splits a whole text line into different parts incorrectly and the training can not converge very well.
	
	\textbf{Influence of the kernel numbers.} We investigate the effect of the number of kernels $n$ on the performance of PSENet. Specifically, we hold the minimal scale $m$ constant and train PSENet with different number of kernels $n$. In details, we set $m$ start from $0.4$ for ICDAR 2015 and $0.6$ for CTW1500 and let $n$ increase from $2$ to $10$. The models are evaluated on ICDAR 2015 and CTW1500 datasets. Fig.~\ref{fig:abs}~(b) shows the experimental results, from which we can find that with the growing of $n$, the F-measure on the test set keeps rising and begins to level off when $n \ge 5$. 
	The advantage of multiple kernels is that it can accurate reconstruct two text instances with large gaps of size where they lying closely to each other.
	
	\textbf{Influence of the backbone.} Deeper neural networks have been proven to boost the performance of large scale image classification and object detection. To better analyze the capability of proposed PSENet, we adopt ResNet as our backbone with three different depths of \{50, 101, 152\} and test on the large scale dataset IC17-MLT. As shown in Table~\ref{tab:backbone}, under the same setting, improving the depth of backbone from 50 to 152 can clearly improve the performance from 70.8\% to 72.2\%, with 1.4\% absolute improvement.
	

	\begin{table}[t]
		\scriptsize
		\centering
		\renewcommand\arraystretch{1}
		\newcommand{\tabincell}[2]{\begin{tabular}{@{}#1@{}}#2\end{tabular}}
		\scalebox{1}{
			\begin{tabular}{|c|c|c|c|}
				\hline
				Methods & P & R & F \\
				\hline
				PSENet~(ResNet50) & 73.7 & 68.2 & 70.8 \\
				\hline
				PSENet~(ResNet101) & 74.8 & 68.9 & 71.7\\
				\hline
				PSENet~(ResNet152) & 75.3	& 69.2	& 72.2\\
				\hline
		\end{tabular}}
		\caption{Performance grows with deeper backbones on IC17-MLT. ``P'', ``R'' and ``F'' represent the precision, recall and F-measure respectively.}
		\label{tab:backbone}
	\end{table}

	\subsection{Comparisons with State-of-the-Art Methods}
	\textbf{Detecting Curve Text}. To test the ability of curve text detection, we evaluate our method on CTW1500 and Total-Text, which mainly contains the curve texts. In the test stage, we scale the longer side of images to 1280 and evaluate the results using the same evaluation method with~\cite{Liu2017Detecting}. 
	We report the single-scale performance of PSENet on CTW1500 and Total-Text in Table~\ref{tab:ctw1500} and Table~\ref{tab:totaltext}, respectively. Note that we only use ResNet50 as the backbone.
	
	On CTW1500, PSENet surpasses all the counterparts even without external data. Notably, we can find that the F-measure (82.2\%) achieved by PSENet is 8.8\% higher than CTD+TLOC and 6.6\% higher than TextSnake on the F-measure. To our best knowledge, this is the best reported result in literature. 
	
	On Total-Text, the proposed PSENet achieves 84.02\%, 77.96\% and 80.87\% in the precision, recall and F-measure, outperforming state-of-the-art methods over 2.47\%. Note that our PSENet extremely surpasses the baseline on Total-Text by more than 40\% in the F-measure. 
	
\begin{figure*}[t]
		\centering
		\setlength{\fboxrule}{0pt}
		\fbox{\includegraphics[width=0.9\textwidth]{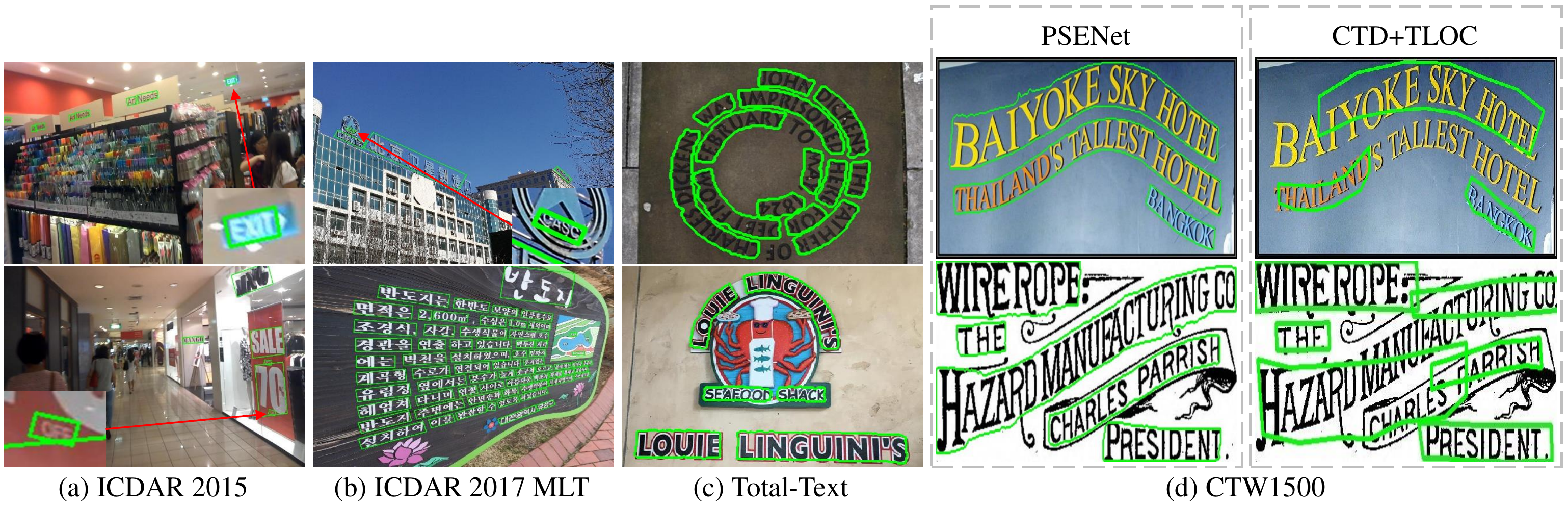}}
		\caption{Detection results on three benchmarks and several representative comparisons of curve texts on CTW1500. More examples are provided in the \textbf{supplementary materials}.}
		\label{fig:res}
	\end{figure*}

	The performance on CTW1500 and Total-Text demonstrates the solid superiority of PSENet when handling curve texts or the texts with arbitrary shapes. We also illustrate several challenging results and make some visual comparisons to the state-of-the-art CTD+TLOC~\cite{Liu2017Detecting} in Fig.~\ref{fig:res}~(d). The comparisons clearly demonstrate that PSENet can elegantly distinguish very complex curve text instances and separate them in a compelling manner. 
	
	\begin{table}[t]
		\scriptsize
		\centering
		\renewcommand\arraystretch{1}
		\newcommand{\tabincell}[3]{\begin{tabular}{@{}#1@{}}#2\end{tabular}}
		\scalebox{1}{
			\begin{tabular}{|c|c|c|c|c|c|}
				\hline
				\multirow{2}{*}{Method} & \multirow{2}{*}{Ext} & \multicolumn{4}{c|}{CTW1500} \\
				\cline{3-6}
				& & P & R & F & FPS\\
				\hline
				CTPN~\cite{tian2016detecting} & - & 60.4* & 53.8* & 56.9* & 7.14\\
				\hline
				SegLink~\cite{shi2017detecting} & - & 42.3* & 40.0* & 40.8* & 10.7 \\
				\hline
				EAST~\cite{zhou2017east} & - & 78.7* & 49.1* & 60.4* & \textbf{21.2} \\
				\hline
				CTD+TLOC~\cite{Liu2017Detecting} & - & 77.4 & 69.8 & 73.4 & 13.3 \\
				\hline
				TextSnake~\cite{textsnake} & \checkmark & 67.9 & 85.3 & 75.6 & - \\
				\hline
				\hline
				PSENet-1s & - &80.6 &75.6 & 78.0 &3.9  \\
				\hline
				PSENet-1s & \checkmark & 84.8 & 79.7 & \textbf{82.2} & 3.9 \\
				\hline
				PSENet-4s & \checkmark & 82.1 & 77.8 & 79.9 & 8.4 \\
				\hline
		\end{tabular}}
		\caption{The single-scale results on CTW1500. ``P'', ``R'' and ``F'' represent the precision, recall and F-measure respectively. ``1s'' and ``4s'' means the width and height of output map is $1/1$ and $1/4$ of the input test image. * indicates the results from~\cite{Liu2017Detecting}. ``Ext'' indicates external data.}
		\label{tab:ctw1500}
	\end{table}

	\begin{table}[t]
		\scriptsize
		\centering
		\renewcommand\arraystretch{1}
		\newcommand{\tabincell}[3]{\begin{tabular}{@{}#1@{}}#2\end{tabular}}
		\scalebox{1}{
			\begin{tabular}{|c|c|c|c|c|c|}
				\hline
				\multirow{2}{*}{Method} & \multirow{2}{*}{Ext} & \multicolumn{4}{c|}{Total-Text} \\
				\cline{3-6}
				& & P & R & F & FPS\\
				\hline
				SegLink~\cite{shi2017detecting} & - & 30.3 & 23.8 & 26.7 & - \\
				\hline
				EAST~\cite{zhou2017east} & - & 50.0 & 36.2 & 42.0 & - \\
				\hline
				DeconvNet~\cite{totaltext} & - & 33.0 & 40.0 & 36.0 & - \\
				\hline
				TextSnake~\cite{textsnake} & \checkmark & 82.7 & 74.5 & 78.4 & - \\
				\hline
				\hline
				PSENet-1s & - & 81.8 & 75.1 & 78.3 & 3.9 \\
				\hline
				PSENet-1s & \checkmark & 84.0 & 78.0 & \textbf{80.9} & 3.9 \\
				\hline
				PSENet-4s & \checkmark & 84.5 & 75.2 & 79.6 & \textbf{8.4} \\
				\hline
		\end{tabular}}
		\caption{The single-scale results on Total-Text. ``P'', ``R'' and ``F'' represent the precision, recall and F-measure respectively. ``1s'' and ``4s'' means the width and height of output map is $1/1$ and $1/4$ of the input test image. ``Ext'' indicates external data. Note that EAST and SegLink were not fine-tuned on Total-Text. Therefore their results are included only for reference.}
		\label{tab:totaltext}
	\end{table}
	
	\textbf{Detecting Oriented Text}. We evaluate the proposed PSENet on the IC15 to test its ability for oriented text detection. Only ResNet50 is adopted as the backbone of PSENet. During inference, we scale the long side of input images to 2240. We compare our method with other state-of-the-art methods in Table~\ref{tab:ic15}. With only single scale setting, our method achieves a F-measure of 85.69\%, surpassing the state of the art results by more than 3\%. In addition, we demonstrate some test examples in Fig.~\ref{fig:res} (a), and PSENet can accurately locate the text instances with various orientations.
	\begin{table}[t]
		\scriptsize
		\centering
		\renewcommand\arraystretch{1}
		\newcommand{\tabincell}[3]{\begin{tabular}{@{}#1@{}}#2\end{tabular}}
		\scalebox{1}{
			\begin{tabular}{|c|c|c|c|c|c|}
				\hline
				\multirow{2}{*}{Method} & \multirow{2}{*}{Ext} & \multicolumn{4}{c|}{IC15} \\
				\cline{3-6}
				& & P & R & F & FPS\\
				\hline
				CTPN~\cite{tian2016detecting} &- & 74.2 & 51.6 & 60.9 & 7.1 \\
				\hline
				SegLink~\cite{shi2017detecting} &\checkmark & 73.1 & 76.8 & 75.0 & - \\
				\hline
				SSTD~\cite{he2017single} &\checkmark & 80.2 & 73.9 & 76.9 & 7.7 \\
				\hline
				WordSup~\cite{hu2017wordsup} &\checkmark & 79.3 & 77.0 & 78.2 & -  \\
				\hline
				EAST~\cite{zhou2017east} & - & 83.6 & 73.5 & 78.2 & \textbf{13.2 }\\
				\hline
				RRPN~\cite{rrpn} &- & 82.0 & 73.0 & 77.0 & - \\
				\hline
				R$^2$CNN~\cite{jiang2017r2cnn} &- & 85.6 & 79.7 & 82.5 & - \\
				\hline
				DeepReg~\cite{deepreg} & - & 82.0 & 80.0 &81.0 & -\\
				\hline
				PixelLink~\cite{PixelLink} & - &82.9 &81.7 &82.3 &7.3\\
				\hline
				Lyu et al.~\cite{lyu2018multi} & \checkmark & 94.1 & 70.7 & 80.7 & 3.6 \\
				\hline
				RRD~\cite{rrd} &\checkmark & 85.6 &79.0 &82.2 &6.5\\
				\hline
				TextSnake~\cite{textsnake} & \checkmark & 84.9 & 80.4 & 82.6 & 1.1 \\
				\hline
				\hline
				PSENet-1s & - & 81.5 & 79.7 & 80.6 & 1.6 \\
				\hline
				PSENet-1s & \checkmark & 86.9 & 84.5 & \textbf{85.7} & 1.6 \\
				\hline
				PSENet-4s & \checkmark & 86.1 & 83.8  & 84.9 & 3.8 \\
				\hline
				
		\end{tabular}}
		\caption{The single-scale results on IC15. ``P'', ``R'' and ``F'' represent the precision, recall and F-measure respectively. ``1s'' and ``4s'' means the width and height of output map is $1/1$ and $1/4$ of the input test image. ``Ext'' indicates external data.}
		\label{tab:ic15}
	\end{table}
	
\textbf{Detecting MultiLingual Text}. To test the robustness of PSENet to multiple languages, we evaluate PSENet on IC17-MLT benchmark. Due to the large scale of the dataset, in order to fully exploit the potential of the PSENet, we adopt Res50 and Res152 as the backbone. We enlarge the original image by 2 times, the proposed PSENet achieve a F-measure of 72.13\%, outperforming state of the art methods by absolute 5.3\%.
In addition, we demonstrate some test examples in Fig.~\ref{fig:res}~(b), and PSENet can accurately locate the text instances with multiple languages.
This proves that PSENet is robust for multi-lingual and multi-oriented detection and can indeed be deployed in complex natural scenarios. 
The result is shown in Table~\ref{tab:ic17}.
	
Note that, We use the high resolution to test IC15 and IC17-MLT because there are so many small texts in these two datasets.
	\begin{table}[t]
		\scriptsize
		\centering
		\renewcommand\arraystretch{1}
		\newcommand{\tabincell}[2]{\begin{tabular}{@{}#1@{}}#2\end{tabular}}
		\scalebox{1}{
			\begin{tabular}{|c|c|c|c|c|c|}
				\hline
				\multirow{2}{*}{Method} & \multirow{2}{*}{Ext} & \multicolumn{3}{c|}{IC17-MLT} \\
				\cline{3-6}
				& & P & R & F  \\
				\hline
				linkage-ER-Flow~\cite{icdar2017mlt} & & 44.48 & 25.59 & 32.49  \\
				\hline
				TH-DL~\cite{icdar2017mlt} & & 67.75 & 34.78 & 45.97 \\
				\hline
				TDN SJTU2017~\cite{icdar2017mlt} & & 64.27 & 47.13 & 54.38 \\
				\hline
				SARI FDU RRPN v1~\cite{icdar2017mlt} & & 71.17 & 55.50 & 62.37 \\
				\hline
				SCUT DLVClab1~\cite{icdar2017mlt} & & 80.28 & 54.54 & 64.96 \\
				\hline
				Lyu et al.~\cite{lyu2018multi} & \checkmark & 83.8 & 55.6 & 66.8 \\
				\hline
				\hline
				PSENet~(ResNet50) & - & 73.77 & 68.21 & 70.88 \\
				\hline
				PSENet~(ResNet152) & - & 75.35 & 69.18 & \textbf{72.13} \\
				\hline
		\end{tabular}}
		\caption{The single-scale results on IC17-MLT. ``P'', ``R'' and ``F'' represent the precision, recall and F-measure respectively. ``Ext'' indicates external data.}
		\label{tab:ic17}
	\end{table}
	
	\subsection{Speed Analyze}
	As shown in Table~\ref{tab:speed}, PSENet can fast detect curve text instance. ResNet50 and ResNet18 are adopted as the backbone to trade off the speed and accuracy. 
	We specially analyze the time consumption of PSENet in different stages. When the output feature map is $1/1$ of the input image, PSENet obtains the best performance, while the time consumption of PSE is more than half of the total inference time because of the larger feature map.
	If the size of output feature map is $1/4$ of the input images, the FPS of PSENet can be boosted from 3.9 to 8.4, while the performance slightly decrease from 82.2\% to 79.9\%, which is shown in Table~\ref{tab:ctw1500}. We can see the time consumption of PSE is less than $1/10$ of total inference time.
	Furthermore, when we scale the long edge of 640, the FPS is further pushed to 22 and the detector still has good performance (75.6\%).
	
	When we use ResNet 18 as the backbone, the speed of PSENet is nearly real-time~(27 FPS), while the performance is still competitive. 
	Note that the PSENet(ResNet18) does not use external data to pretrain.
	Combined with Table~\ref{tab:ctw1500}, we can find PSENet surpasses EAST and CTD+TLOC in both speed and performance.

	All of the above experiments are tested on CTW1500 test set. We evaluate all test images and calculate the average speed. We scale the long edge of \{1280, 960, 640\} as input to test the speed. All results in Table~\ref{tab:speed} are tested by PyTorch~\cite{pytorch} and one 1080Ti GPU.

	\begin{table}[t]
		\scriptsize
		\centering
		\renewcommand\arraystretch{1}
		\newcommand{\tabincell}[3]{\begin{tabular}{@{}#1@{}}#2\end{tabular}}
		\scalebox{0.8}{
			\begin{tabular}{|c|c|c|c|c|c|c|c|}
				\hline
				\multirow{2}{*}{Method} & \multirow{2}{*}{Res} & \multirow{2}{*}{F} & \multicolumn{3}{c|}{Time consumption} & \multirow{2}{*}{FPS} \\
				\cline{4-6}
				& & & backbone(ms) & head(ms) & PSE(ms) &\\
				\hline
				PSENet-1s~(ResNet50) & 1280 & 82.2 & 50 & 68 & 145 & 3.9\\
				\hline
				PSENet-4s~(ResNet50) & 1280 & 79.9 & 50 & 60 & 10 & 8.4 \\
				\hline
				PSENet-4s~(ResNet50) & 960 & 78.3 & 33 & 35 & 9 & 13 \\
				\hline
				PSENet-4s~(ResNet50) & 640 & 75.6 & 18 & 20 & 8 & 21.65 \\
				\hline
				\hline
				PSENet-4s$^\dagger$~(ResNet18) & 960 & 74.3 & 10 & 17 & 10 & 26.75 \\
				\hline
		\end{tabular}}
		\caption{Time consumption of PSENet on CTW-1500. The total time is consist of backbone, head of segmentation and PSE part. $\dagger$ indicates training from scratch. ``Res'' represents the resolution of the input image. ``F'' represent the F-measure.}
		\label{tab:speed}
	\end{table}

	\section{Conclusion and Future Work}
	We propose a novel Progressive Scale Expansion Network (PSENet) to successfully detect the text instances with arbitrary shapes in the natural scene images. By gradually expanding the detected areas from small kernels to large and complete instances via multiple semantic segmentation maps, our method is robust to shapes and can easily separate those text instances which are very close or even partially intersected. The experiments on scene text detection benchmarks demonstrate the superior performance of the proposed method. 
	
	There are multiple directions to explore in the future. Firstly, we will investigate whether the expansion algorithm can be trained along with the network end-to-end. Secondly, the progressive scale expansion algorithm can be introduced to the general instance-level segmentation tasks, especially in those benchmarks with many crowded object instances. We are cleaning our codes and will release them soon.
	
	\section{Acknowledgment}
	This work is supported by the Natural Science Foundation of China under Grant 61672273 and Grant 61832008, the Science Foundation for Distinguished Young Scholars of Jiangsu under Grant BK20160021, and Scientific Foundation of State Grid Corporation of China (Research on Ice-wind Disaster Feature Recognition and Prediction by Few-shot Machine Learning in Transmission Lines).

\section{Supplementary Materials for Shape Robust Text Detection with Progressive Scale Expansion Network}

\subsection{The Advantage of Multiple Kernels}
\begin{figure}[t]
		\centering
		\setlength{\fboxrule}{0pt}
		\fbox{\includegraphics[width=0.45\textwidth]{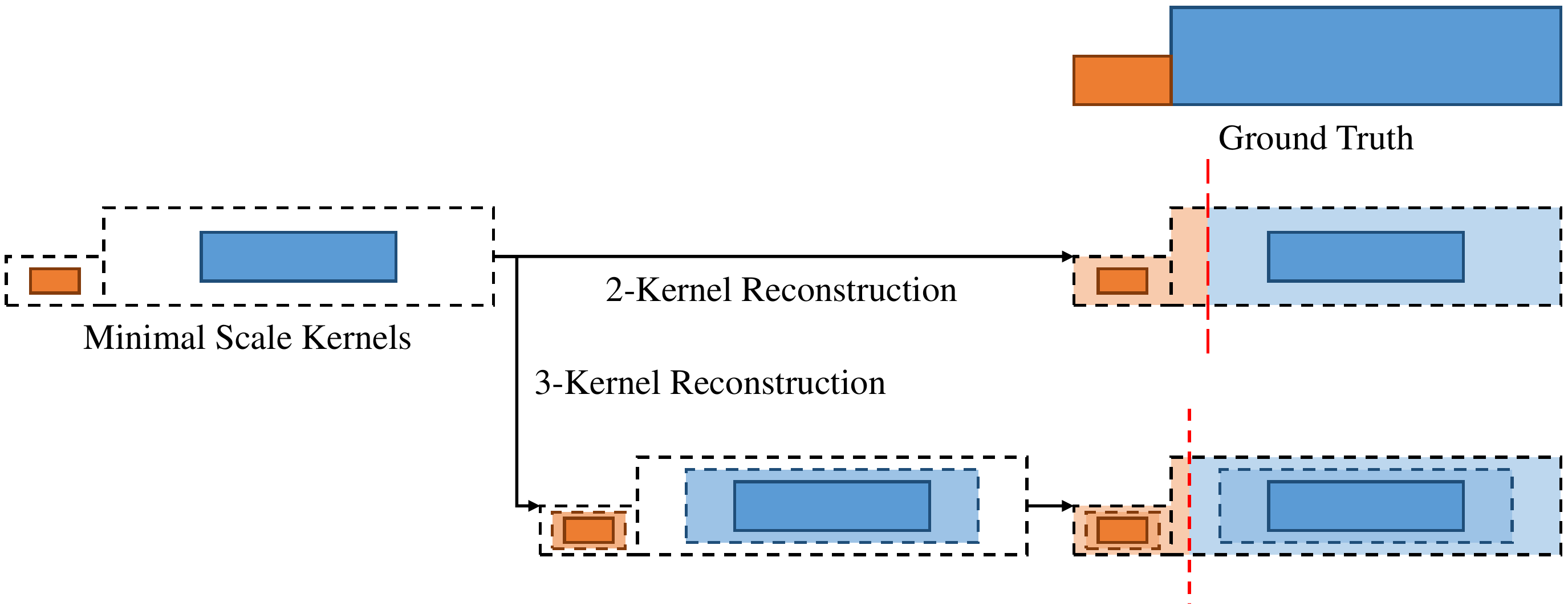}}
		\caption{The difference of 2-kernel reconstruction and 3-kernel reconstruction.}
		\label{fig:kernel_num_fig}
	\end{figure}

	Multiple kernels is used to reconstruct the closely text instances smoothly 
	when their sizes have large gap. As shown in Fig.~\ref{fig:kernel_num_fig}, there are some flaws in 2-kerenl reconstruction, and this problem is alleviated when the number of kernels increasing (see 3-kernel reconstruction).
	In addition, the time complexity of progressive scale expansion algorithm (PSE) is $\mathcal{O}(W \times H)$\footnote{To reduce time complexity, we set the end pixels of the $i$th expansion as the start pixels of the $(i+1)$th expansion.}, where $W \times H$ is the size of output.
	Thus, the increasement of kernel number have no influence on the time cost of PSE. Consequently, it is a good manner to use multiple kernels to reconstruct the text instances.
	
	\subsection{Applying PSENet on Other Semantic Segmentation Framework}
	The proposed PSENet consists of two key points: kernel mechanism and PSE. Both of them are easy to be applied on other semantic segmentation frameworks. Here, we implement PSENet-like method based on a widely used semantic segmentation framework PSPNet~\cite{zhao2017pyramid}, and evaluate it on CTW1500. We detailly compare the PSENet-like method based on PSPNet with the original PSENet in Table \ref{tab:speed_psp}. We can find the method based on PSPNet can also achieve competitive performance on the curve text dataset. However, compared with the original PSENet, the PSPNet-based method need more GPU memory (3.7G vs 2.9G) and have lower forward speed (289ms vs 118ms), which indicates that the original PSENet is more suitable to text detection.
	
	\subsection{More Comparisons on CTW1500}
\begin{figure*}[b]
		\centering
		\setlength{\fboxrule}{0pt}
		\fbox{\includegraphics[width=0.75\textwidth]{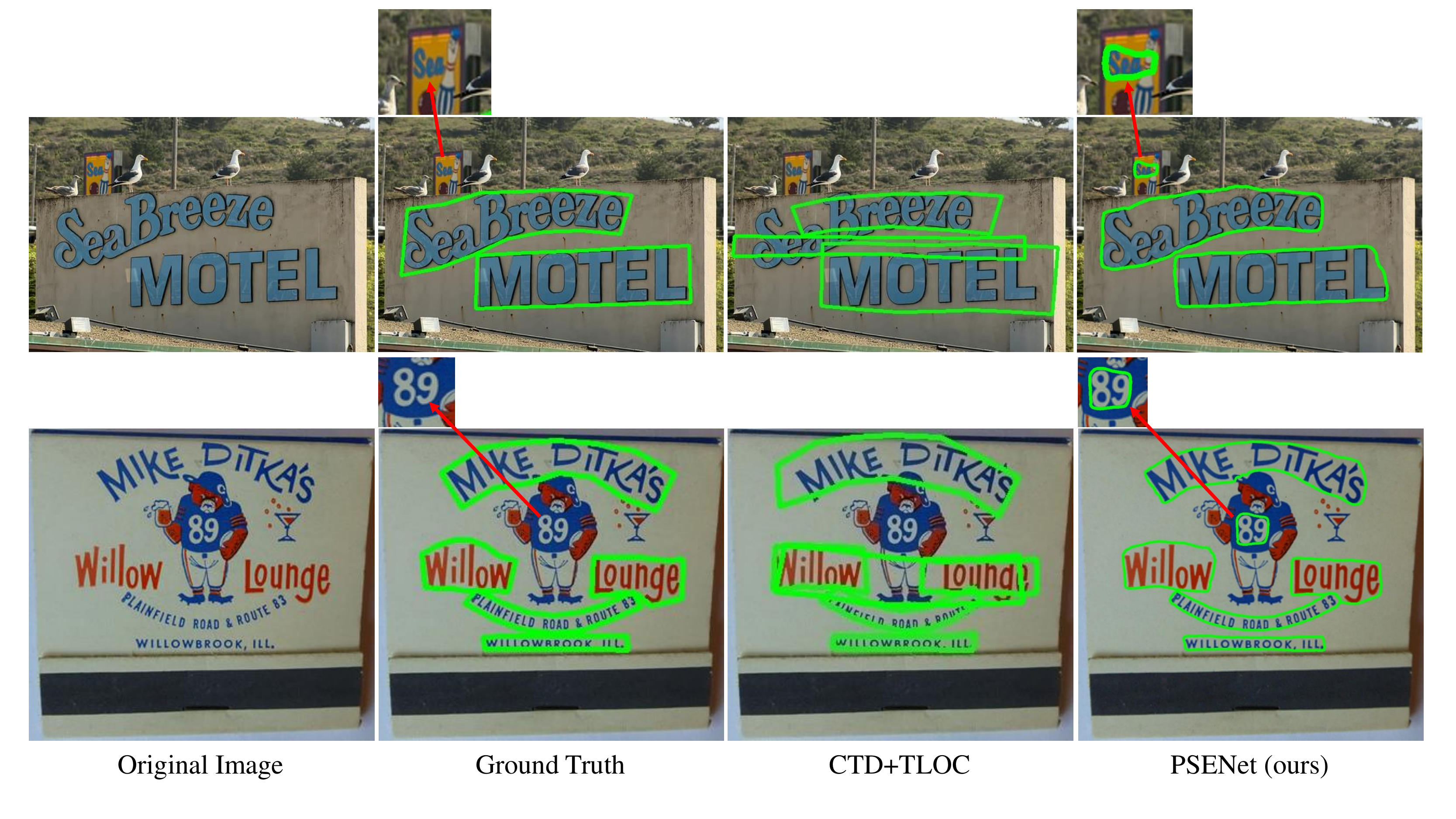}}
		\caption{Comparisons on CTW1500. The proposed PSENet produces several detections that are even missed by the groundtruth labels.}
		\label{fig:cmp_ctw1}
	\end{figure*}

\begin{figure*}[b]
		\centering
		\setlength{\fboxrule}{0pt}
		\fbox{\includegraphics[width=0.75\textwidth]{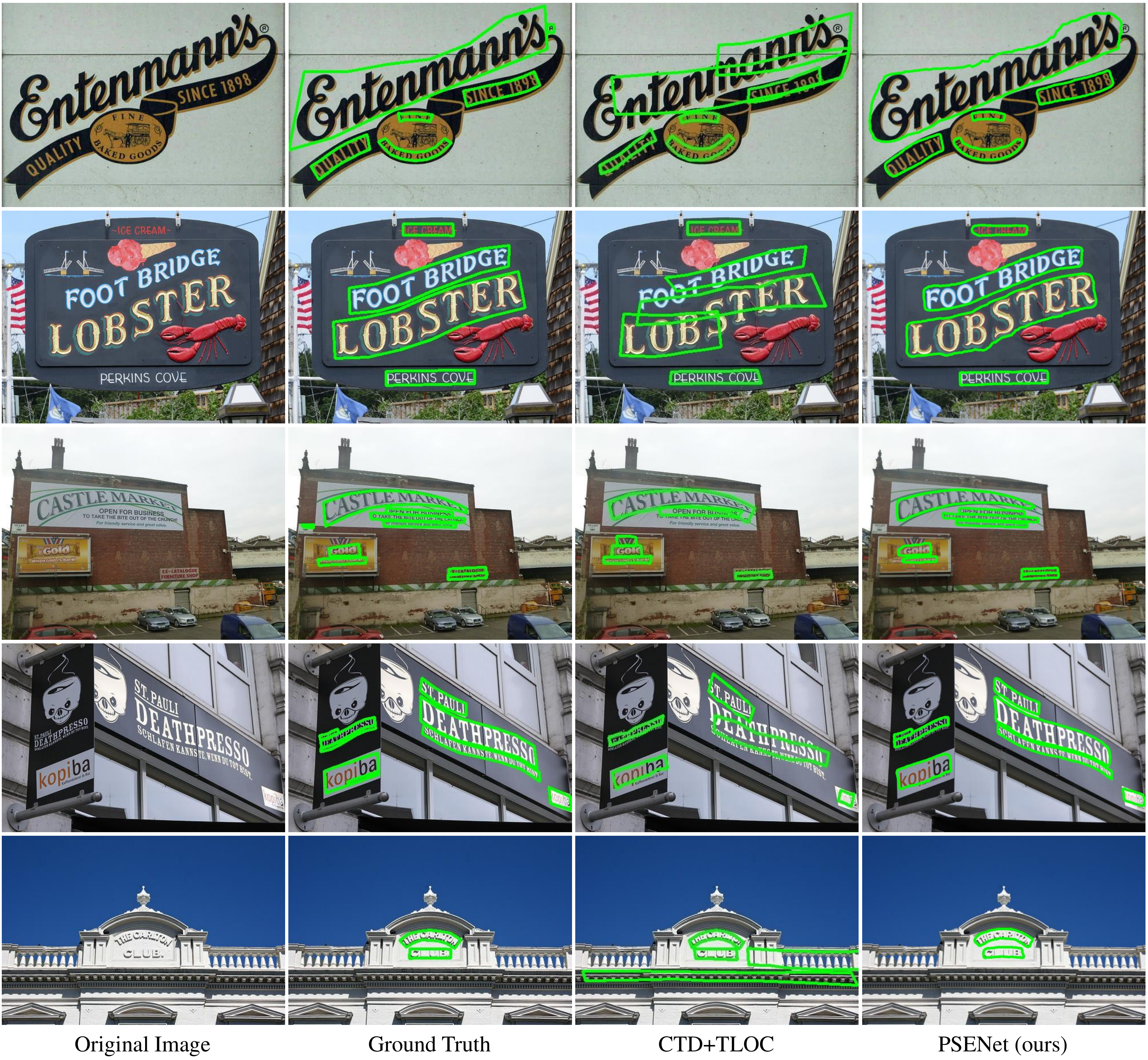}}
		\caption{Comparisons on CTW1500.}
		\label{fig:cmp_ctw2}
	\end{figure*}

	To demonstrate the power of PSENet on complex curve text detection, we implement the state-of-the-art and open source~\footnote{https://github.com/Yuliang-Liu/Curve-Text-Detector} method CTD-TLOC, and make detail comparisons between PSENet and CTD-TLOC on CTW1500.
	The comparisons are shown in Fig.~\ref{fig:cmp_ctw1},~\ref{fig:cmp_ctw2}. 
	It is interesting and amazing to find that in Fig.~\ref{fig:cmp_ctw1}, our proposed PSENet is able to locate several text instances where the groundtruth labels are even unmarked. This highly proves that our method is quite robust due to its strong learning representation and distinguishing ability. Fig.~\ref{fig:cmp_ctw2} demonstrate more examples where PSENet can not only detect the curve text instances even with extreme curvature, but also separate those close text instances in a good manner.
	
	\begin{table}[t]
		\scriptsize
		\centering
		\renewcommand\arraystretch{1}
		\newcommand{\tabincell}[3]{\begin{tabular}{@{}#1@{}}#2\end{tabular}}
		\scalebox{0.8}{
			\begin{tabular}{|c|c|c|c|c|c|c|}
				\hline
				\multirow{2}{*}{Method} & \multirow{2}{*}{Mem (G)} & \multirow{2}{*}{F} & \multicolumn{2}{c|}{Time consumption (ms)} & \multirow{2}{*}{FPS} \\
				\cline{4-5}
				& & & Forward & PSE &\\
				\hline
				PSENet-1s~(ResNet50) & 2.9 & 77.98 & 118 & 145 & 3.9\\
				\hline
				PSPNet~\cite{zhao2017pyramid}~+~PSE-1s~(ResNet50) & 3.7 & 77.25 & 289 & 145 & 2.3\\
				\hline
		\end{tabular}}
		\caption{Time consumption of PSENet-like method based on PSPNet and original PSENet. Both of them are trained from scratch. Mem means GPU memory. F means F-measure. “1s” means the size of output map is equal to input image.}
		\label{tab:speed_psp}
	\end{table}

	
	\subsection{More Detected Examples on Total Text, ICDAR 2015 and ICDAR 2017 MLT}
\begin{figure*}[b]
		\centering
		\setlength{\fboxrule}{0pt}
		\fbox{\includegraphics[width=0.75\textwidth]{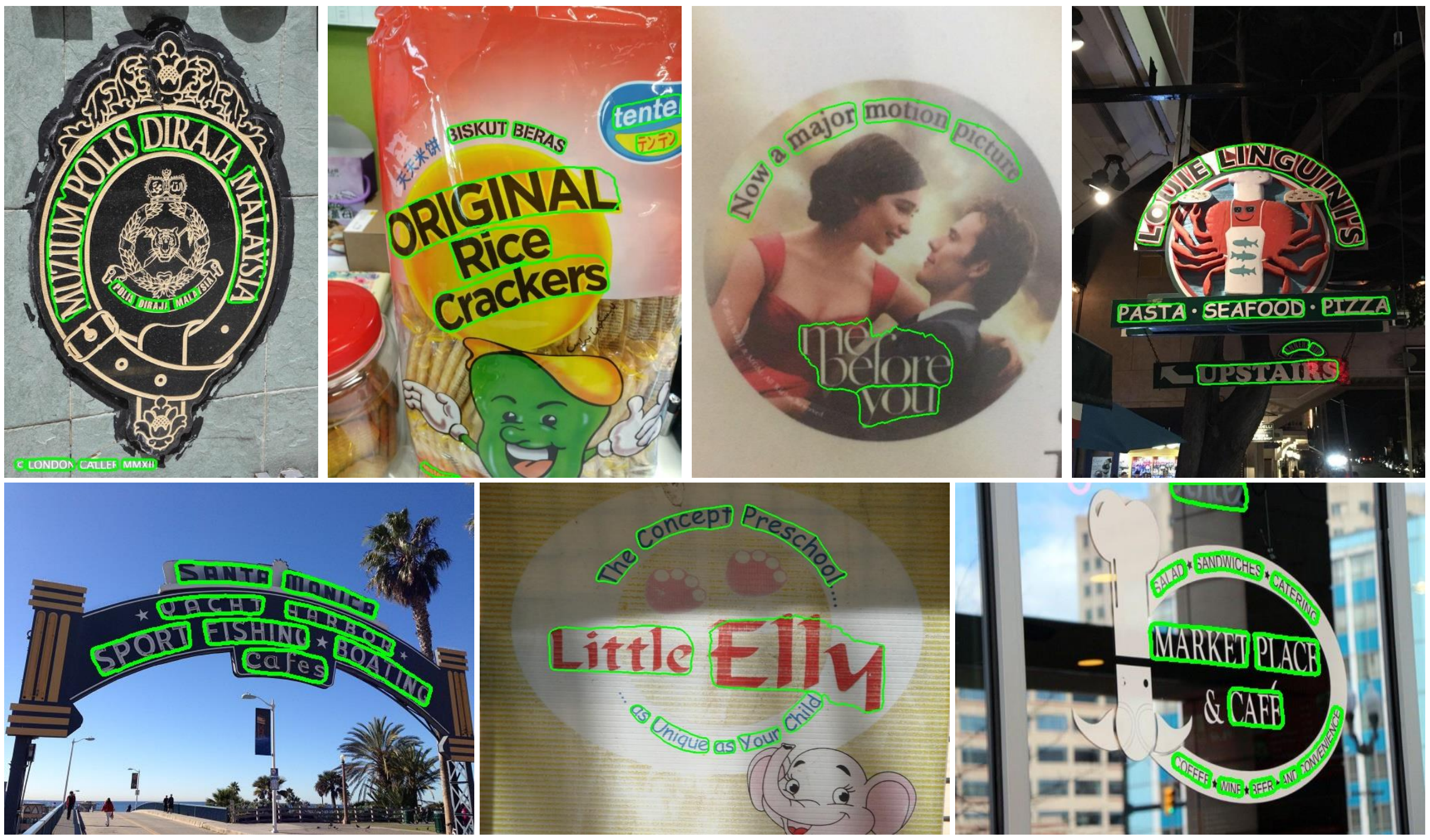}}
		\caption{Test examples on Total Text produced by PSENet.}
		\label{fig:tt}
	\end{figure*}

\begin{figure*}[b]
		\centering
		\setlength{\fboxrule}{0pt}
		\fbox{\includegraphics[width=0.75\textwidth]{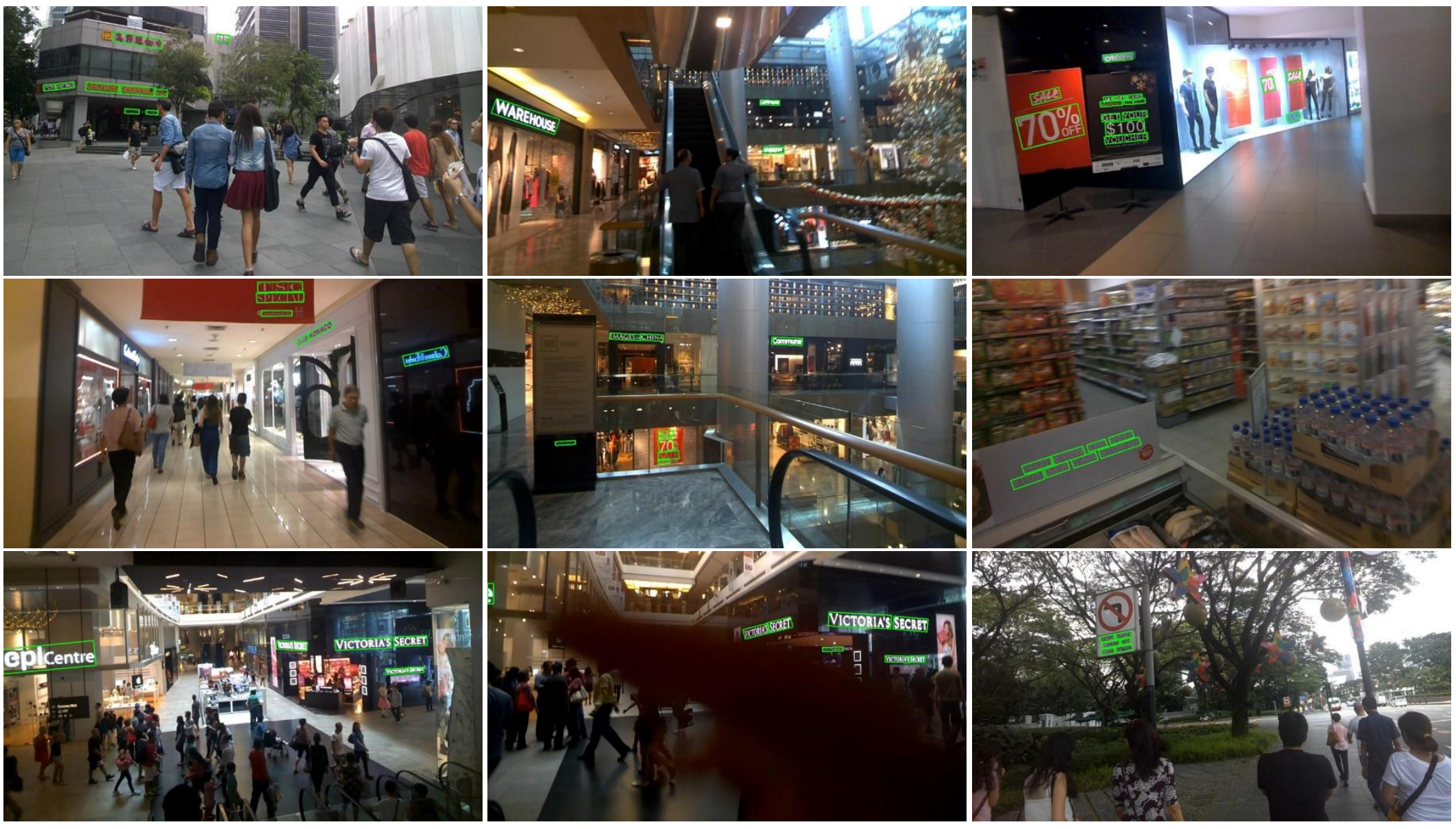}}
		\caption{Test examples on ICDAR 2015 produced by PSENet.}
		\label{fig:ic15}
	\end{figure*}

\begin{figure*}[b]
		\centering
		\setlength{\fboxrule}{0pt}
		\fbox{\includegraphics[width=0.75\textwidth]{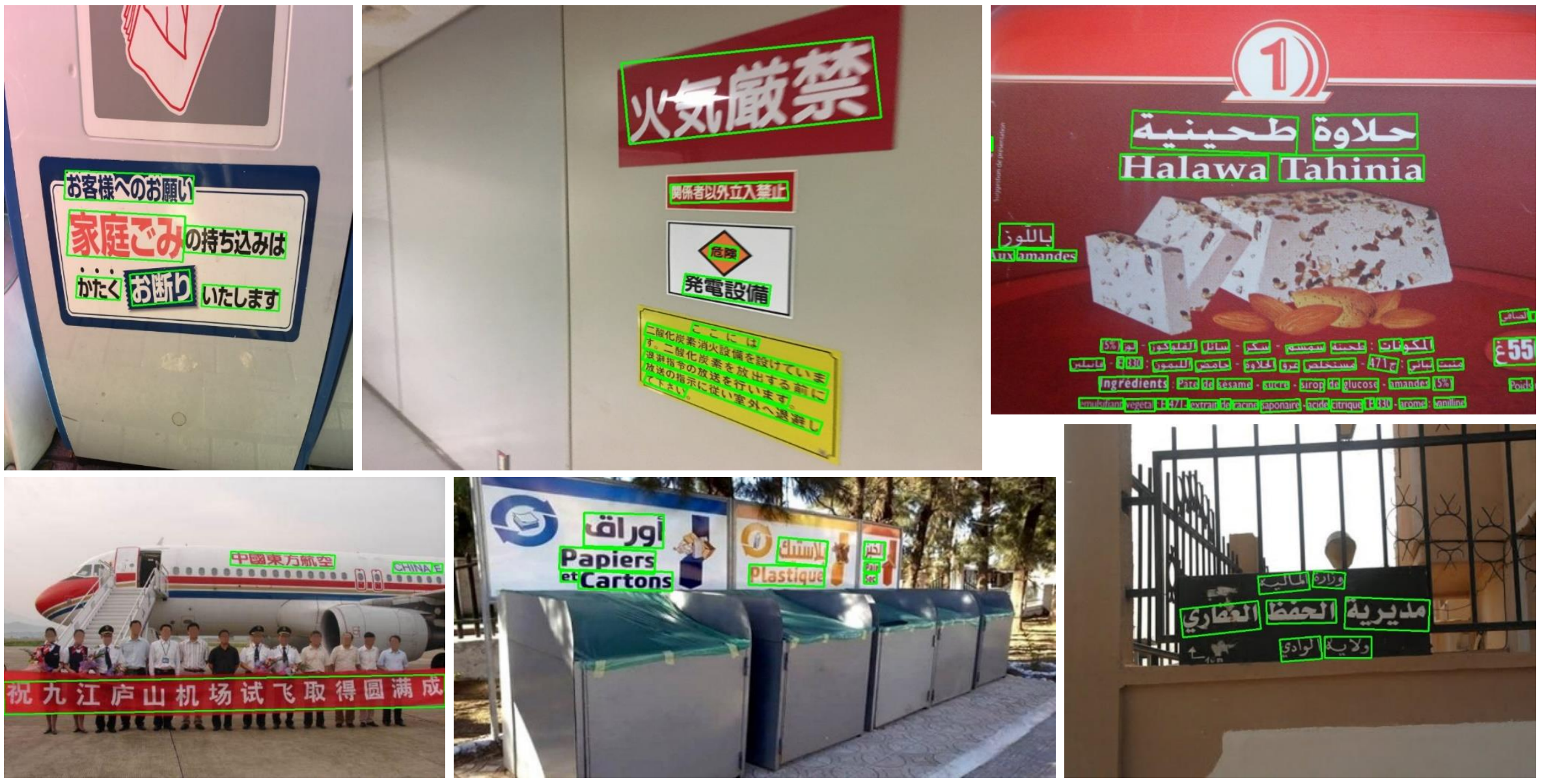}}
		\caption{Test examples on ICDAR 2017 MLT produced by PSENet.}
		\label{fig:ic17}
	\end{figure*}

	In this section, we demonstrate more test examples produced by the proposed PSENet in Fig.~\ref{fig:tt} (Total Text), Fig.~\ref{fig:ic15}
	(ICDAR 2015) and Fig.~\ref{fig:ic17} (ICDAR 2017 MLT).
	From these results, it can be easily observed that with the proposed kernel-based framework and PSE, our method is able to archive the following points:
	1) locating the arbitrary-shaped text instances precisely;
	2) separating the closely adjacent text instances well;
	3) detecting the text instances with various orientations;
	4) detecting the multiLingual text.
	Meanwhile, thanks to the strong feature representation, PSENet can as well locate the text instances with complex and unstable illumination, different colors and variable scales.

{\small
\bibliographystyle{ieee}
\bibliography{egbib}
}

\end{document}